\documentclass{article}

\PassOptionsToPackage{numbers, compress}{natbib}

\usepackage[preprint]{neurips_2024}

\usepackage[utf8]{inputenc} 
\usepackage[T1]{fontenc}    
\usepackage{hyperref}       
\usepackage{url}            
\usepackage{booktabs}       
\usepackage{amsfonts}       
\usepackage{nicefrac}       
\usepackage{microtype}      
\usepackage{xcolor}         
\usepackage{graphicx}
\usepackage{amsmath}
\usepackage{amssymb}
\usepackage{algorithm}
\usepackage{algpseudocode}
\usepackage{enumitem}
\usepackage{verbatim}

\newcommand{\ours}{\textbf{CoNo}}
\newcommand{\egno}{\textit{e.g.}}
\newcommand{\ieno}{\textit{i.e.}}
\usepackage{marvosym}
\title{CoNo: Consistency Noise Injection for Tuning-free Long Video Diffusion}

\author{%
  Xingrui Wang, Xin Li\textsuperscript{~\Letter} and Zhibo Chen\textsuperscript{~\Letter} \\
  University of Science and Technology of China\\
  \texttt{\small{\{wxrui\_18264819595,lixin666\}@mail.ustc.edu.cn, chenzhibo@ustc.edu.cn}} \\
  }

\begin{document}

\maketitle
\renewcommand{\thefootnote}{}
\footnotetext{$^{\textrm{\Letter}}$  Corresponding authors.}

\begin{abstract}
Tuning-free long video diffusion has been proposed to generate extended-duration videos with enriched content by reusing the knowledge from pre-trained short video diffusion model without retraining. However, most works overlook the fine-grained long-term video consistency modeling, resulting in limited scene consistency (\textit{i.e.}, unreasonable object or background transitions), especially with multiple text inputs. To mitigate this, we propose the \textbf{Co}nsistency \textbf{No}ise Injection, dubbed~\ours, which introduces the ``look-back'' mechanism to enhance the fine-grained scene transition between different video clips, and designs the long-term consistency regularization to eliminate the content shifts when extending video contents through noise prediction. In particular, the ``look-back'' mechanism breaks the noise scheduling process into three essential parts, where one internal noise prediction part is injected into two video-extending parts, intending to achieve a fine-grained transition between two video clips. The long-term consistency regularization focuses on explicitly minimizing the pixel-wise distance between the predicted noises of the extended video clip and the original one, thereby preventing abrupt scene transitions. Extensive experiments have shown the effectiveness of the above strategies by performing long-video generation under both single- and multi-text prompt conditions. The project has been available in \url{https://wxrui182.github.io/CoNo.github.io/}.

\end{abstract}

\section{Introduction}
The fast advancement of artificial intelligent generative contents (AIGCs) has revolutionized the way humans create and interact. Meanwhile, video generation, as the representative direction of AIGCs, is significantly promoted from the perspectives of perceptual quality~\cite{singer2022make-a-video,ho2022imagen,ho2022vdm}, lengths~\cite{he2022lvdm,gu2023ReuseAndDiffuse}, and customization~\cite{chen2023videodreamer,chen2023MCDiff,wei2023dreamvideo,xing2024MakeYourVideo,ma2024magic}, by excavating the advantages of diffusion models and large collected video datasets from the internet, such as SoRA~\cite{Brooks2024Sora}. In particular, text-to-video generation~\cite{wang2023lavie,guo2023animatediff,zhuang2024vlogger} has demonstrated significant potential in interactive creation, \egno, short-form UGC video creation and movie production, by generating video content consistent with provided text prompts, thereby attracting considerate attention. Despite that, limited by unaffordable training resources and imperfect video representation learning, recent text-to-image video generation still suffers from unsatisfied video frame length and scene consistency~\cite{wang2023modelscope,zhang2023show}. It is urgent to develop text-to-video generation methods that can produce long and scene-consistent video content with fewer resources.

To eliminate the above challenges, several studies have been proposed to achieve long video generation. The first category~\cite{brooks2022generating,Brooks2024Sora} regards the video as a whole for representation learning and extends the frame lengths by increasing the computational costs. The second one empowers the generation models with the frame interpolation/extrapolation capability through training, which generates partial frames first and then extends them in an autoregressive or hierarchical manner~\cite{villegas2022phenaki,ge2022TATS,he2022lvdm,chen2023seine,wang2024magicvideo,yin2023nuwaxl}. However, the above methods still rely on extensive training and struggle to adapt to multiple text prompts. In contrast, tuning-free long video generation, as a new paradigm, is proposed to reuse the off-the-shelf pre-trained video generation models to extend the video frames and seamlessly achieve scene transitions with multiple text prompts. The whole process does not need additional training resources, \egno, extensive data, and GPU costs,  which is applicable to multiple base video generation modules that meet different user needs~\cite{wang2023gen-l-video,qiu2023freenoise}.

In this work, we focus on tuning-free long video generation and propose a novel long video generation method, \ieno,  Consistency Noise Injection (\ours) to eliminate the primary limitations in existing tuning-free works:  (i) Coarse transition between different video clips. Existing works~\cite{wang2023gen-l-video,oh2023mtvg} typically utilize simple latent/attention fusion mechanisms to ensure a consistent transition with the last few frames of the previously generated video clip. (ii) Overlooking explicit long-term content consistency modeling. Most works~\cite{qiu2023freenoise,gu2023ReuseAndDiffuse} achieve content consistency through the inherent implicit consistency modeling capability of pre-trained generation models.

Our \ours~solves the first limitation by introducing the ``look-back'' mechanism. In contrast to previous works that directly extend video clips guided with one-side frames, the ``look-back'' mechanism divides the video extension process into three crucial stages, where one internal noise prediction stage is inserted into two video extending stages, intending to ensure stable content transition through the inherent constrain of two-side contents at each reverse process (\ieno, the predicted noises from left existing predicted frames and right extending frames). Notably maintaining the overall initial noise group of different video clips is crucial to guarantee the same contents/scenes~\cite{qiu2023freenoise,ge2023preserve}, we also propose customized noise shuffle strategies for the above three stages, respectively. Concretely, we design the revised extending noise shuffle for the video extending stage, which recovers the noise order for guided frames after reversing the whole initial noises, thereby obviating the reverse-order repetitive content generation. For internal noise prediction, we directly inserted the initial noises at the end of the sequence into the middle position, resulting in an internal noise shuffle to ensure the same initial noise group. To solve the second limitation, we propose an explicit long-term consistency regularization, which minimizes the pixel-wise distance between predicted noises of the extended video clip and the original generated video clip. The purpose is to reduce the possibility of scene/content shifts occurring when extending the videos. With the above two innovations, our \ours~achieves  
state-of-the-art scene consistency and perceptual quality on tuning-free long video generation under both single- and multi-text prompt conditions.

The contributions of this paper are summarized as follows:
\begin{itemize}[leftmargin=*]
    \item We propose a brand-new tuning-free long video diffusion with our proposed consistency noise injection, intending to enhance the fine-grained long-term consistency of generated long videos.
    \item Our \ours~is composed of two innovations: (i) the proposed ``look-back" mechanism achieves the fine-grained scene transition between different video clips by incorporating the internal noise prediction and two customized noise shuffle strategies, and (ii) the proposed long-term consistency regularization is used to eliminate the content shifts occurred in extended videos.
    \item Extensive experiments under both single- and multi-text prompt conditions have demonstrated the effectiveness of our approach with extensive results.
\end{itemize}

\section{Related Work}
\label{sec:related work}
\paragraph{Text-to-Video Generation.} Text-to-video generation aims to transform textual descriptions into semantically aligned videos. Initially, most video generation models, primarily those based on GANs~\cite{li2018video,kim2020tivgan} or transformers~\cite{yan2021videogpt,wu2021godiva}, were limited to more restricted datasets. In contrast, current diffusion-based models~\cite{ho2022vdm,singer2022make-a-video,ho2022imagen,wang2023modelscope,qing2023hierarchical,bar2024lumiere,chen2024videocrafter2,girdhar2023emu,luo2023videofusion} demonstrate advanced capabilities in creating realistic scenarios. Within this category, Stable Diffusion (SD) based methods garner significant attention for their efficiency and exceptional performance~\cite{blattmann2023VideoLDM,chen2023videocrafter1,he2022lvdm,wang2023lavie,guo2023animatediff,khachatryan2023Text2VideoZero,wu2023freeinit,an2023latent,xing2023simda}. These methods exploit the generative power of pre-trained image diffusion models~\cite{rombach2022StableDiffusion,zhang2023ControlNet} and enhance their functionality by integrating temporal modules. Additionally, a novel subgroup of diffusion-based models integrates transformer blocks~\cite{vaswani2017attention} to model temporal relationships by segmenting videos into spatial-temporal patches, effectively utilizing the sequence modeling capabilities of transformers~\cite{lu2023vdt,chen2023gentron,gao2024lumina}. Diffusion-based methods have become mainstream and continue to inspire a wide range of customized applications~\cite{feng2023ccedit,guo2023sparsectrl,ma2024FollowYourPose,yin2023dragnuwa,wu2023Next-gpt,liu2023GenerativeDisco,zhang2023i2vgen,hu2023lamd,ni2023conditional,wang2024videocomposer}.
\paragraph{Long Video Generation.} The generation of long videos is increasingly attracting attention due to its promising applications and unique challenges. One approach requires significant computational resources for training~\cite{villegas2022phenaki,harvey2022FDM,henschel2024streamingt2v}. These methods either generate new content autoregressively~\cite{ge2022TATS,yu2023PVDM,harvey2022FDM,voleti2022mcvd} or employ a coarse-to-fine approach, sampling keyframes and then interpolating additional frames~\cite{yin2023nuwaxl}. Notably, the Sora model~\cite{Brooks2024Sora} stands at the forefront of these computationally intensive methods, capable of producing a minute of high-fidelity video. To reduce computational costs, another category adopts tuning-free methods that utilize existing short video models for resource-friendly extensions. For instance, Gen-L-Video~\cite{wang2023gen-l-video} inferences longer videos and maintains content consistency through temporal co-denoising across multiple short videos. FreeNoise~\cite{qiu2023freenoise} improves consistency using noise rescheduling and sliding window-based attention fusion. Moreover, MTVG~\cite{oh2023mtvg} leverages an inversion technique for initial latent codes that apply to new prompts. However, most previous studies exhibit more divergent denoising trajectories when generating new content, as they impose only coarse consistency constraints to the initial noise or the locally overlapping latent codes, thus leading to scene inconsistency. In this work, we propose a ``look-back'' mechanism to enhance consistency more finely from both side contents and correct the denoising direction at each timestep through long-term consistency regularization.
\section{Methodology}
\label{Methodology}
Our method consists of two key components, described in Sec.~\ref{sec: content-guided denoising} and Sec.~\ref{sec:the look-back mechanism} respectively. In Sec.~\ref{sec: content-guided denoising}, we begin by presenting an observation, which leads to the introduction of long-term consistency regularization to avoid abrupt content shifts. Sec.~\ref{sec:the look-back mechanism} details the ``look-back'' mechanism designed to enhance fine-grained scene consistency. The preliminaries related to the aforementioned sections are introduced in Sec.~\ref{sec:Preliminaries}.
\begin{figure}[!ht]
    \centering
    \includegraphics[width=\textwidth]{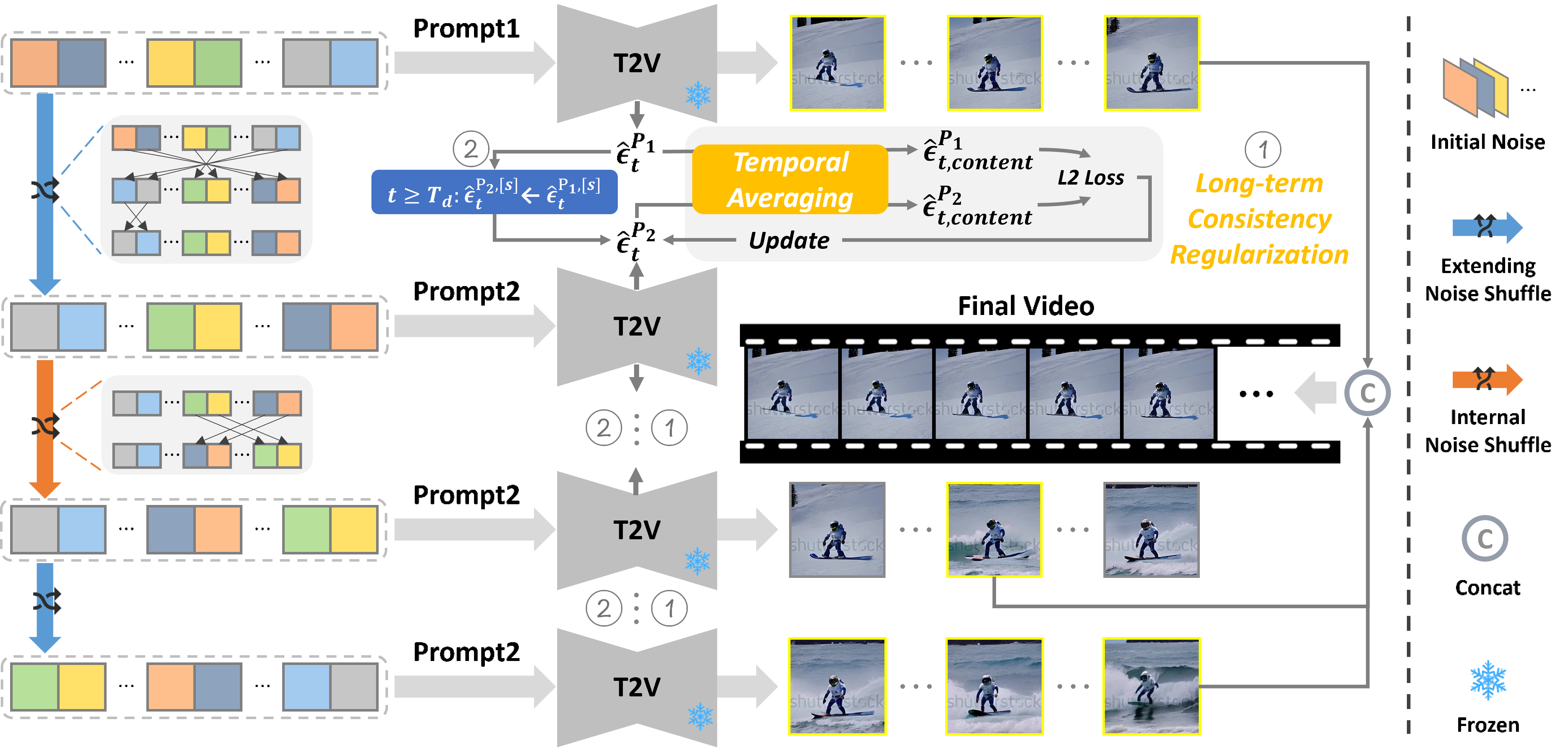}
    \caption{\textbf{Illustration of the CoNo framework.} We propose a ``look-back'' mechanism that inserts an internal noise prediction stage between two video extending stages to enhance scene consistency. To achieve this, we design the extending and internal initial noise shuffles and constrain the denoising trajectory using selected predicted noise (denoted as $\left[s\right]$ in the figure). Additionally, we apply long-term consistency regularization between adjacent video clips to avoid abrupt content shifts. We obtain the final video by concatenating the frames marked with yellow boxes from different stages.}
    \label{fig:framework}
\end{figure}
\vspace{-5mm}
\subsection{Preliminaries}
\label{sec:Preliminaries}
\textbf{Diffusion Models}~\cite{ho2020ddpm, song2020ddim, song2020ScoreBased} (DMs) incrementally disrupt the data distribution, $x_0\sim q\left( x_0 \right)$, by introducing Gaussian noise through a process known as \textit{diffusion}, which comprises $T$ timesteps:
\begin{equation}
    q\left( x_{1:T}|x_0 \right) =\prod_{t=1}^T{q\left( x_t|x_{t-1} \right),  \quad \quad  q\left( x_t|x_{t-1} \right) =\mathcal{N} \left( x_t|\sqrt{\alpha _t}x_{t-1},\beta _tI \right)}
\end{equation}
with $t$ denoting the timestep, $\beta_t$ as a predefined variance schedule, and $\alpha_t=1-\beta_t$. In the \textit{denoising} process of diffusion, a model denoted by $\epsilon_{\theta}$ and parameterized by $\theta$ is trained to predict the noise, enabling the iterative recovery of $x_0$ from $x_T$:
\begin{equation}
\underset{\theta}{\min}\mathbb{E} _{t,x_0,\epsilon \sim \mathcal{N} \left( 0,\mathrm{I} \right)}\left\| \epsilon -\epsilon _{\theta}\left( x_t;c,t \right) \right\| _{2}^{2},
\end{equation}
where $\epsilon$ represents the noise injected into $x_0$ to obtain $x_t$, and $c$ denotes an optional conditioning signal, such as a text prompt.

\textbf{Latent Diffusion Model} (LDM)~\cite{rombach2022StableDiffusion} was proposed to reduce computational and memory resources by executing the generation process in the latent space. Given a sample $x_0\in \mathbb{R} ^{3\times H\times W}$, it is mapped to the latent code $z_0\in \mathbb{R} ^{c\times h\times w}$ by a regularized autoencoder that employs an encoder $\mathcal{E}$ and a decoder $\mathcal{D}$:
\begin{equation}
z_0=\mathcal{E} \left( x_0 \right) , \quad \quad \hat{x}_0=\mathcal{D} \left( z_0 \right) .
\end{equation}
Consequently, the \textit{diffusion} and \textit{denoising} processes are applied within this learned low-dimensional latent space:
\begin{equation}
\hat{\epsilon}=\epsilon _{\theta}\left( z_t;c,t \right)
\end{equation}

\textbf{Video Latent Diffusion Models} (VideoLDMs)~\cite{blattmann2023VideoLDM, ge2023PYoCo, gu2023ReuseAndDiffuse} extend the LDM framework with temporal modules to enable text-to-video generation. By integrating temporal attention and convolution layers into off-the-shelf pre-trained image LDMs and subsequently fine-tuning on videos, these image diffusion models are transformed into capable video generators. The latent representation $z_0$, now extending into a four-dimensional space $\mathbb{R}^{c\times n\times h\times w}$ with additional frames $n$, endows the model parameters $\theta$ with temporal awareness.

\begin{figure}[!ht]
    \centering
    \includegraphics[width=0.99\linewidth]
    {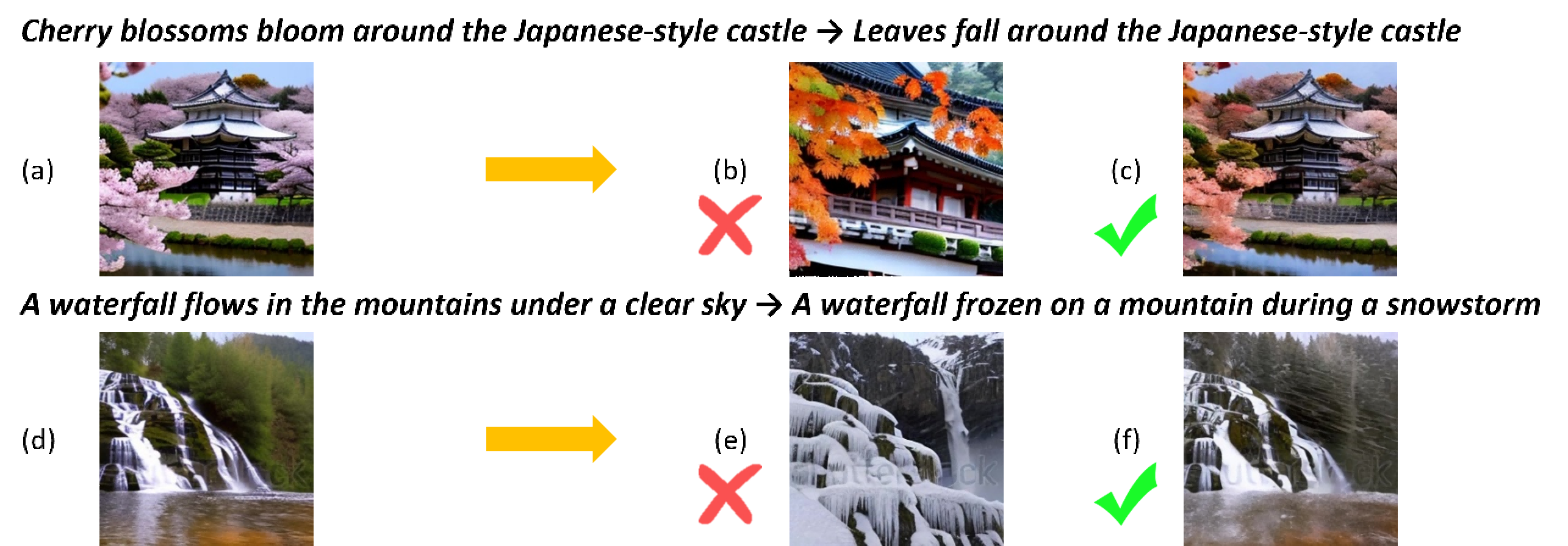}
    \caption{Observed content shifts and improvements brought by the proposed Long-term Consistency Regularization.}
    \label{fig:with_grad}
\end{figure}
\vspace{-5mm}
\subsection{Observation and Long-term Consistency Regularization}
\label{sec: content-guided denoising}
\paragraph{Observation.} Maintaining the content consistency of the generated video clips is a key issue, whereas even with the same initial noise, the content generated under different text prompts exhibits significant divergence, as demonstrated in Fig.~\ref{fig:with_grad}. This discrepancy is illustrated by comparing Figs.~\ref{fig:with_grad} (a) and (b), where the scenario described in (a) is ``Cherry blossoms bloom around the Japanese-style castle,'' versus (b) ``Leaves fall around the Japanese-style castle.'' 
By employing the pre-trained VideoCrafter1~\cite{chen2023videocrafter1} model to generate videos under identical initial noise conditions (showcasing only the first frame for clarity), we observe significant changes in the appearance of elements like cherry blossoms and the Japanese-style castle in the video frame. The phenomenon is further exemplified by the examples shown in Figs.~\ref{fig:with_grad} (d) and (e). Naturally, a pivotal question arises: How to eliminate the content shifts between video clips with different text prompts?

\vspace{-3mm}
\paragraph{Long-term Consistency Regularization.} To address this question, we focus our attention on the \textit{denoising} process and propose \textit{Long-term Consistency Regularization}.
It's evident that the latent code $z_0 \in \mathbb{R}^{c\times n\times h\times w}$ for a video clip of $N$ frames and the iteratively generated noise $\hat{\epsilon} \in \mathbb{R}^{c\times n\times h\times w}$ both encompass the motion and content dimensions, which motivates us to minimize the long-term pixel-wise distance of predicted noise between different clips. 
We use $\hat{\epsilon}_t$ to represent the predicted noise at timestep $t$. 
For two distinct text prompts $P_1$ and $P_2$ with the same group of initial noise, we derive $\hat{\epsilon}_{t}^{P_1}$ and $\hat{\epsilon}_{t}^{P_2}$. 
Aiming to make the content of extended video clip (inputting $P_2$) consistent with the originally generated video clip (inputting $P_1$), we define an $L2$ loss function
\begin{equation}
    \scalebox{0.85}{$ 
        g\left( \hat{\epsilon}_{t,content}^{P_1},\hat{\epsilon}_{t,content}^{P_2} \right) =\left\| \frac{\sum_{n=0}^{N-1}{\left( \hat{\epsilon}_{t}^{P_1}-\hat{\epsilon}_{t}^{P_2} \right)}}{N} \right\| _{2}^{2}=\left\| \frac{\sum_{n=0}^{N-1}{\hat{\epsilon}_{t}^{P_1}}}{N}-\frac{\sum_{n=0}^{N-1}{\hat{\epsilon}_{t}^{P_2}}}{N} \right\| _{2}^{2}=\left\| \hat{\epsilon}_{t,content}^{P_1}-\hat{\epsilon}_{t,content}^{P_2} \right\| _{2}^{2}
    $}
\end{equation}
to compare content at each step $t$, and subsequently update the current predicted noise $\hat{\epsilon}_{t}^{P_2}$ in the direction that minimizes $g$:
\begin{equation}
    \hat{\epsilon}_{t}^{P_2}\gets \hat{\epsilon}_{t}^{P_2}-\delta \nabla _{\hat{\epsilon}_{t}^{P_2}}g\left( \hat{\epsilon}_{t,content}^{P_1},\hat{\epsilon}_{t,content}^{P_2} \right), 
\end{equation}
where $\delta$ is a scalar that determines the step size of the update. 
The $L2$ loss function is established on the $N$-frame scale of the base video diffusion model, seeking to make the predicted noise of corresponding frames from different video clips as close as possible, so that the overall denoising trajectories of both clips are gradually unifying. Employing this regularization to regulate the \textit{denoising} path in a long-term manner ensures the content of the generated videos remains consistent across varying text prompts, as illustrated in Figs.~\ref{fig:with_grad} (c) and (f).
\subsection{The ``Look-Back'' Mechanism for Video Clip Transition}
\label{sec:the look-back mechanism}
\vspace{-2mm}
Long-term Consistency Regularization establishes a robust foundation for further video clip transition. While we have achieved content consistency among different video clips, resolving transitions between distinct scenes remains a challenge. In this section, we leverage the existing priors of video diffusion models to tackle this. Pre-trained on video datasets, video diffusion models exhibit notable temporal perception capabilities, indicating that, with suitable guidance, these models can effectively execute both video frame internal and external sampling. To verify this conjecture, we continue to explore from the perspective of noise and further propose the ``look-back'' mechanism that consists of three stages, iteratively looking back at the already sampled contents to generate reasonable extending or internal transitions.
\begin{figure}[!h]
    \centering
    \includegraphics[width=0.99\linewidth]
    {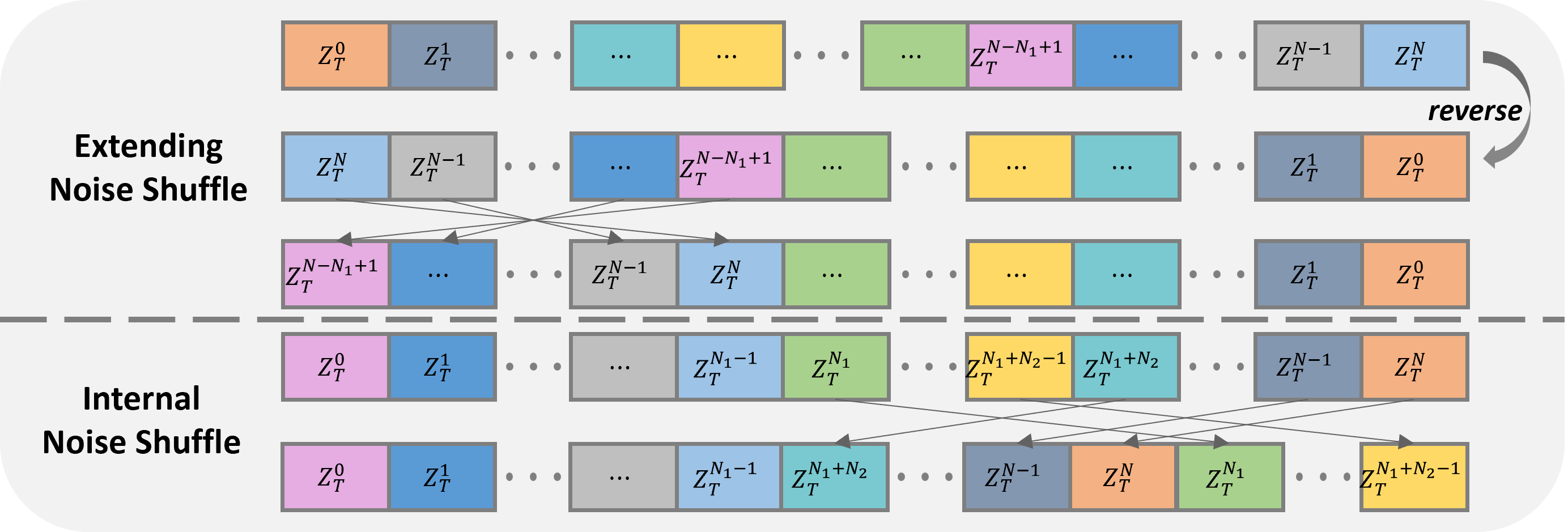}
    \caption{To constrain the denoising trajectory of selected frames, we design extending and internal noise shuffles for the initial noise. Different colored blocks represent different video frames, with $z_T$ indicating the initial noise and frame numbers annotated in the top right corner.}
    \label{fig:ExtraAndInter}
\end{figure}
\vspace{-5mm}
\paragraph{Video Extending Stage.} The video extending stage in the ``look-back'' mechanism is designed to extend the current scene. For any given text prompt $P_1$, we initially sample a set of noise $z_{T}^{P_1}\sim \mathcal{N} \left( 0,I \right)$ and generate the first video clip $z_{0}^{P_1}$ without imposing any constraints. 
The recent study~\cite{qiu2023freenoise} has revealed maintaining the same group of initial noise (even randomly shuffled) is vital to ensure consistent scenes between video clips, which inspires us to set the initial noise for various video clips as $z_{T}^{P_1}$, but with customized noise shuffle strategies. Therefore, we developed the \textit{Extending Noise Shuffle} for this video extending stage. Specifically, when generating the following video clip $z_{0}^{P_2}$, we first reverse $z_{T}^{P_1}$, and then further reverse its first $N_1$ frames, that is:
\begin{equation}
\begin{split}
    z_{T}^{P_2,i}=z_{T}^{P_1,N-i-1}, \quad \quad i\in \left\{ 0,1,\cdots ,N-1 \right\}, 
\\
z_{T}^{P_2,i}=z_{T}^{P_2,N_1-i-1}, \quad \quad i\in \left\{ 0,1,\cdots ,N_1-1 \right\},
\end{split}
\end{equation}
which is illustrated in Fig.~\ref{fig:ExtraAndInter} (a series of numbers is assigned to illustrate the changes in order). After reordering the initial noise, we ensure that $z_{T}^{P_1,\left\{ N-N_1,\cdots ,N-1 \right\}}$ copies $z_{T}^{P_2 ,\left\{ 0,\cdots ,N_1-1 \right\}}$, enabling us to constrain the denoising trajectories on the initial $N_1$ frames of $z_{T}^{P_2}$ that ensures its fully denoised frames are consistent with the last $N_1$ frames of $z_{0}^{P_1}$. We first store the predicted noise $\hat{\epsilon}_{t}^{P_1}$ at each timestep $t$ during the gradual denoising to obtain $z_{0}^{P_1}$. When denoising $z_{T}^{P_2}$ under the guidance of text prompt $P_2$, the predicted noise $\hat{\epsilon}_{t}^{P_2,\left\{ 0,\cdots ,N_1-1 \right \} }$ will be replaced by $\hat{\epsilon}_{t}^{P_1,\left\{ N-N_1,\cdots ,N-1 \right\}}$ before the predefined timestep $T_d$. Because $z_{ T} ^{P_1,\left\{ N-N_1,\cdots ,N-1 \right\}}=z_{T}^{P_2,\left\{ 0,\cdots ,N_1-1 \right \} }$ and $\hat{\epsilon}_{t\geqslant T_d}^{P_1,\left\{ N-N_1,\cdots ,N-1 \right\}}=\hat{\epsilon}_{t\geqslant T_d}^{P_2,\left\{ 0,\cdots ,N_1-1 \right\}}$, the denoising trajectories of these $N_1$ frames are almost the same, thus $z_{0}^{P_1,\left\{ N-N_1,\cdots ,N-1 \right\}}\approx z_{0}^{P_2,\left\{ 0,\cdots ,N_1-1 \right\}}$ and the remaining $ N-N_1$ frames are sampled using the priors of the base video diffusion model. These initial $N_1$ frames play a role as guidance, which ensures the extending frames inherit the previous scene. We can arbitrarily choose whether $P_1$ equals $P_2$ to facilitate video extending transition under single- or multi-text prompt conditions. 
\vspace{-3mm}
\paragraph{Internal Noise Prediction Stage.} After the video extending stage, $z_{0}^{P_2,\left\{ N_1,\cdots ,N-1 \right\}}$ now reflects the semantics of text prompt $P_2$. To enhance the stable content transition between various scenes $z_{ 0}^{P_2,\left\{ 0,\cdots ,N_1-1 \right\}}$ and $z_{0}^{P_2,\left\{ N_1,\cdots ,N-1 \right\}}$, we utilize the inherent constraints of above two-side contents for internal noise prediction. We accordingly designed the \textit{Internal Noise Shuffle} strategy. Particularly, we select $N_2$ initial noise frames, namely $z_{T}^{P_2,\left\{ N_1,\cdots ,N_1+N_2-1 \right\}}$, and reposition them at the end of the noise sequence. The frames $z_{T}^{P_2,\left\{ 0,\cdots ,N_1-1 \right\}}$ remain unchanged, while the rest, $N-N_1-N_2$ noise frames, are shifted to occupy the space between these two segments, ensuring the initial noise remains the same set. After that, we perform internal noise prediction under the guidance of text prompt $P_2$ and represent the predicted noise at each step $t$ as $\hat{\epsilon}_{t}^{P_{2^{\prime}}}$. Now we have $z_{T}^{P_{2^{\prime}},\left\{ 0,\cdots ,N_1-1 \right\}}=z_{T}^{P_2,\left\{ 0,\cdots ,N_1-1 \right\}}$ and $z_{T}^{P_{2^{\prime}},\left\{ N-N_2,\cdots ,N-1 \right\}}=z_{T}^{P_2,\left\{ N_1,\cdots ,N_1+N_2-1 \right\}}$. By substituting $\hat{\epsilon}_{t\geqslant T_d}^{P_{2^{\prime}},\left\{ 0,\cdots ,N_1-1 \right\}}$ and $\hat{\epsilon}_{t\geqslant T_d}^{P_{2^{\prime}},\left\{ N-N_2,\cdots ,N-1 \right\}}$ with $\hat{\epsilon}_{t\geqslant T_d}^{P_2,\left\{ 0,\cdots ,N_1-1 \right\}}$ and $\hat{\epsilon}_{t\geqslant T_d}^{P_2,\left\{ N_1,\cdots ,N_1+N_2-1 \right\}}$ before the timestep $T_d$ respectively, we constrain the denoising trajectories at both ends, enabling the intermediate $N-N_1-N_2$ transition frames inference.
Compared to the extending transition stage solely relying on one-side frames, the internal noise prediction, strongly constrained by both the left existing frames and the right extending frames, promotes more appropriate scene transitions, as demonstrated in Sec.~\ref{sec:ablation}.

To further increase the video length of scene $P_2$, after obtaining $z_{0}^{P_{2^{\prime}}}$, we use $P_2$ to perform the video extending stage again, resulting in $z_{0}^{P_{2^{''}}}$. It is notable that long-term consistency regularization, which is discussed in Sec.~\ref{sec: content-guided denoising}, is applied between adjacent video clips during both video extending and internal noise prediction stages, and its absence results in compromised scene consistency, as further explored in Sec.~\ref{sec:ablation}. 

We show the complete procedure in Fig.~\ref{fig:framework}, where long-term consistency regularization precedes the noise replacement step of the ``look-back'' mechanism. We should clarify that the first video extending stage serves as a precursor to introducing new scenes, and the video frames obtained from this stage are not used in the final output. Ultimately, we concatenate the initial video clip $z_{0}^{P_1}$, the transition frames $z_{0}^{P_{2^{\prime}},\left\{ N_1,\cdots ,N-N_1-1 \right\}}$, and $z_{0}^{P_{2^{''}}}$ to generate the extended video. When the prompts $P_1$ and $P_2$ are the same, we maintain continuity within the same scene, whereas different text prompts allow for transitions between various scenes. This video expansion process is iteratively performed by inputting additional prompts while the video length gradually increases. Compared to MTVG~\cite{oh2023mtvg}, which only uses the last frame of the previous video clip to constrain consistency, we consider richer temporal information to achieve better video quality. Furthermore, we explore the improvements in video continuity brought by prompt engineering in appendix~\ref{sec: extended experiments} and provide the pseudo-code of CoNo in appendix~\ref{sec:pseudo-code}.

\section{Experiments}
\label{sec: Experiments}
\subsection{Experiment Setup}
\label{sec:set_up}
\vspace{-3mm}
\paragraph{Implementation Details.} We conducted experiments using the open-source video generation model, VideoCrafter1~\cite{chen2023videocrafter1}. VideoCrafter1 is trained to generate short videos consisting of $16$ frames at a resolution of $256\times 256$. To evaluate the model’s performance with a single text prompt, we used Evalcrafter~\cite{liu2023evalcrafter}, and for multiple text prompts, we primarily utilized the test set from MTVG~\cite{oh2023mtvg}. Under single-text conditions, we expanded the video twice, with corresponding comparison models generating $64$ frames. Meanwhile, under multi-text conditions, the number of expansions was adjusted according to the number of input texts. We also conducted experiments with Lavie~\cite{wang2023lavie} to validate the generalization of CoNo, as detailed in Sec.~\ref{sec: Generalization Validation}. All experiments were performed using a single NVIDIA GeForce RTX $3090$.
\vspace{-3mm}
\paragraph{Evaluation Metrics.} Following prior works~\cite{qiu2023freenoise,oh2023mtvg}, we calculate the FVD~\cite{unterthiner2018FVD_KVD} and KVD~\cite{unterthiner2018FVD_KVD} between the original short videos and segments of equivalent lengths derived from extended videos. We report the CLIP-Image score~\cite{radford2021ClipSim,wu2023tune-a-video} to evaluate the semantic similarity between two consecutive frames, assessing content consistency. Moreover, the CLIP-Text score~\cite{hessel2021clipscore} is used to measure the alignment between the given text prompts and the generated video frames.
\vspace{-3mm}
\paragraph{Compared Methods.} To validate the effectiveness of CoNo, we compare it with several state-of-the-art (SOTA) tuning-free methods. When inputting a single text prompt, we employ Gen-L-Video (GenL)~\cite{wang2023gen-l-video} and FreeNoise~\cite{qiu2023freenoise} for inference. In scenarios conditioned on multiple text prompts, we evaluate the results using GenL~\cite{wang2023gen-l-video}, VidRD~\cite{gu2023ReuseAndDiffuse}, FreeNoise~\cite{qiu2023freenoise}, MTVG~\cite{oh2023mtvg}, and our model.
\subsection{Single-prompt Longer Video Generation}
\label{sec:single-prompt}
\begin{table}[!h]
    \footnotesize
    \centering
    \caption{Quantitative comparison of single-prompt longer video generation.}
    \begin{tabular}{l|ccc|ccc}
\toprule
\multicolumn{1}{l}{} & \multicolumn{3}{c}{Automatic Metric} & \multicolumn{3}{c}{Human Evaluation}                                                                                   \\ \midrule
Method               & FVD$\downarrow$       & KVD$\downarrow$      & CLIP-Image$\uparrow$    & \multicolumn{1}{l}{Semantic$\uparrow$} & \multicolumn{1}{l}{Temporal$\uparrow$} & \multicolumn{1}{l}{Preference$\uparrow$} \\ \midrule
GenL~\cite{wang2023gen-l-video}   & 177.63    & 21.06    & 0.9370        & 3.41     & 1.96       & 2.26                                  \\
FreeNoise~\cite{qiu2023freenoise} & \underline{85.83}     & \underline{7.06}     & \textbf{0.9732}     & \underline{3.50}       & \underline{3.14}  & \underline{3.04} \\
Ours                 & \textbf{54.13}     & \textbf{1.02}     & \underline{0.9725}        & \textbf{3.78}       & \textbf{3.50}  & \textbf{3.47}                                  \\ \bottomrule
\end{tabular}
    \label{tab:single_long}
\end{table}
\begin{figure}[!h]
    \centering
    \includegraphics[width=0.99\linewidth]
    {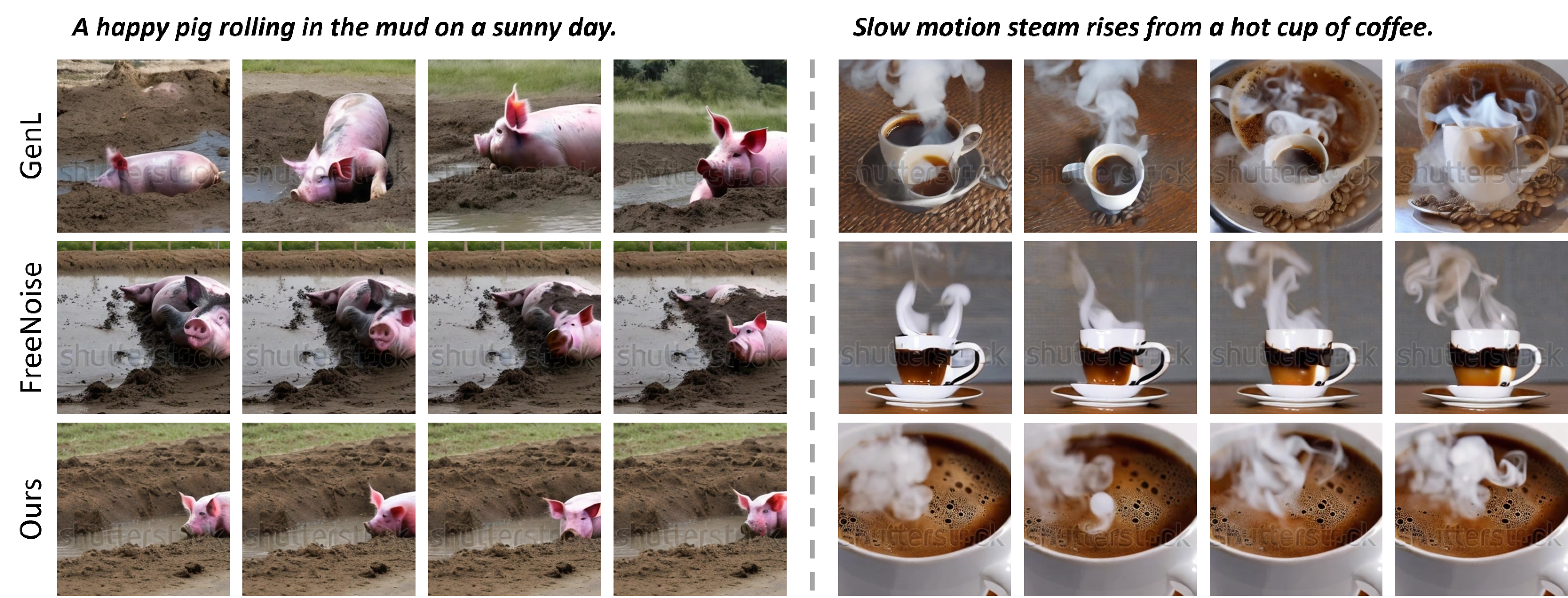}
    \caption{Qualitative comparisons of single-prompt longer video generation.}
    \label{fig:single_long}
\end{figure}
Fig.~\ref{fig:single_long} illustrates the qualitative comparisons given a single text prompt. GenL~\cite{wang2023gen-l-video} and FreeNoise~\cite{qiu2023freenoise} employ a sliding window technique, with GenL’s coarse averaging of overlapping latent codes from adjacent windows inducing mutations in both scene foreground and background. FreeNoise generates visually smoother results due to local window-based attention fusion, yet it still exhibits slow scene changes, such as the gradual enlargement of a coffee cup, which leads to inconsistencies. Benefiting from Long-term Consistency Regularization, our method enhances the content consistency of objects and achieves better video quality. More qualitative results can be found in the appendix~\ref{sec: More Qualitative Results}.

We show quantitative results in Tab.~\ref{tab:single_long}. Through the evaluation procedure described in Sec.~\ref{sec:set_up}, we obtained the best FVD and KVD, while remaining competitive in CLIP-Image score compared to other models. The SOTA results are highlighted in bold, and the second-best results are underlined. We find that the segments sliced from long videos generated by CoNo not only closely match the distribution of those produced by the base model but also maintain frame-to-frame consistency. Furthermore, a user study was conducted to assess users' evaluations of different models in terms of Video-Text Alignment (Semantic), Content Consistency (Temporal), and Video Quality (Preference). We design a questionnaire following the five-point scale from MTVG~\cite{oh2023mtvg} to score the three mentioned aspects, where higher scores indicate better model performance. We provide a detailed description of the human evaluation in the appendix~\ref{sec: Additional Implementation Details}.
\begin{figure}[!h]
    \centering
    \includegraphics[width=0.99\linewidth]
    {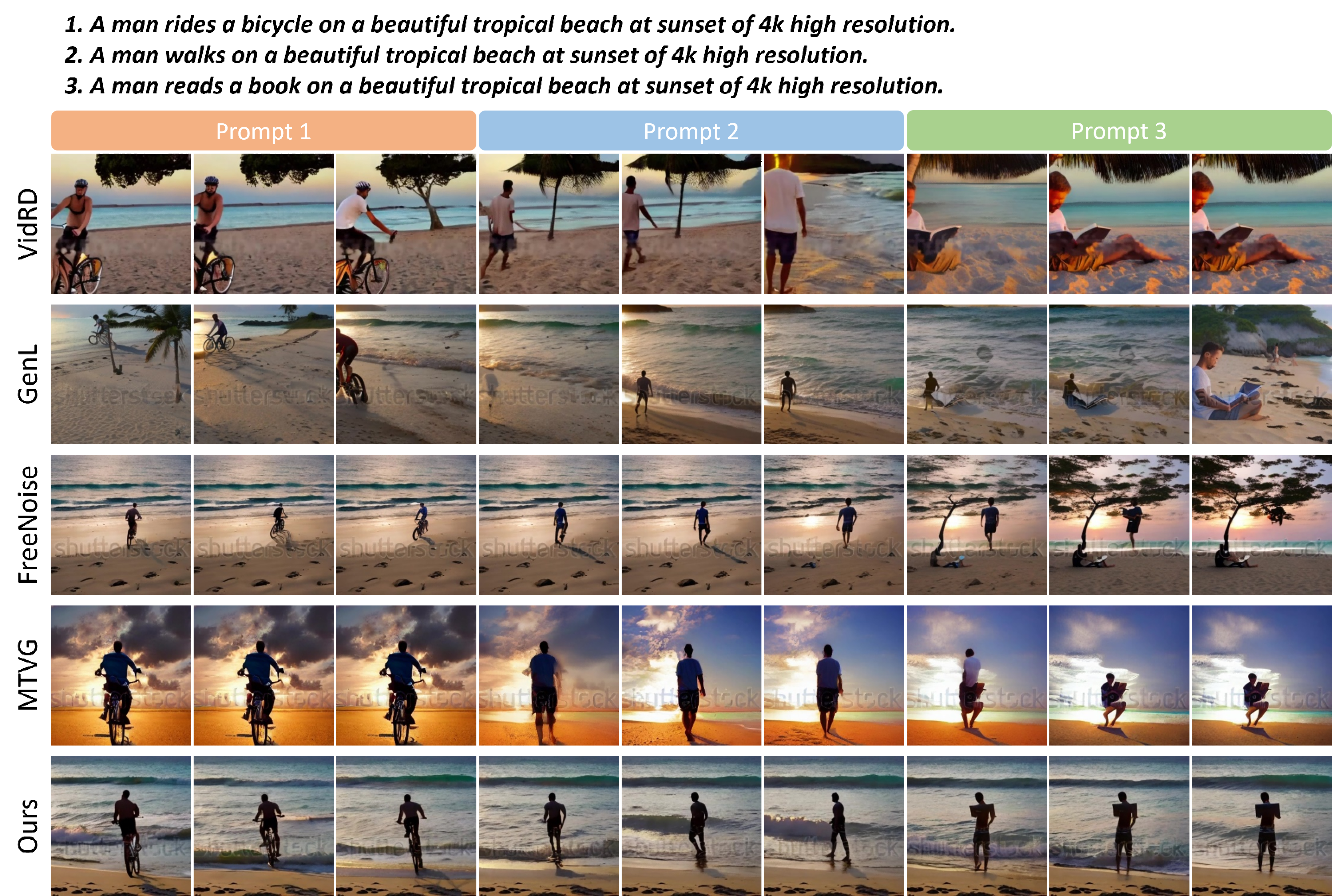}
    \caption{Qualitative comparisons of multi-prompt longer video generation.}
    \label{fig:multi_long}
\end{figure}
\begin{table}[!h]
    \footnotesize
    \centering
    \caption{Quantitative comparison of multi-prompt longer video generation.}
    \begin{tabular}{l|cc|cccc}
\toprule
\multicolumn{1}{l}{} & \multicolumn{2}{c}{Automatic Metric} & \multicolumn{4}{c}{Human Evaluation}                                                                                   \\ \midrule
Method               & CLIP-Text$\uparrow$  & CLIP-Image$\uparrow$    & \multicolumn{1}{l}{Semantic$\uparrow$} & \multicolumn{1}{l}{Temporal$\uparrow$} & \multicolumn{1}{l}{Realism$\uparrow$}  & \multicolumn{1}{l}{Preference$\uparrow$} \\ \midrule
GenL~\cite{wang2023gen-l-video}    & 0.308     & 0.953     & 3.15    & 2.09   & 1.81 & 2.13 \\
VidRD~\cite{gu2023ReuseAndDiffuse} & 0.287     & 0.951     & 3.12    & 2.11   & 2.03 & 2.20 \\
MTVG~\cite{oh2023mtvg}             & 0.309     & 0.957     & \textcolor{gray}{\textbf{3.47}}    & \textcolor{gray}{2.72}   & \textcolor{gray}{2.41} & \textcolor{gray}{\underline{2.70}} \\
FreeNoise~\cite{qiu2023freenoise}  & \underline{0.325}     & \textbf{0.974}     & 3.04    & \underline{3.08}   & \underline{2.43} & 2.67  \\
Ours             & \textbf{0.326}     & \underline{0.967}     & \underline{3.43} & \textbf{3.43} & \textbf{2.71} & \textbf{2.84}    \\
\bottomrule
\end{tabular}
    \label{tab:multi_long}
\end{table}
\subsection{Multi-prompt Longer Video Generation}
\label{sec:multi-prompt}
\vspace{-3mm}
CoNo also handles scenarios with a sequence of different text prompts effectively. We present an example in Fig.~\ref{fig:multi_long}, and more videos are provided in the appendix~\ref{sec: More Qualitative Results}. As depicted in Fig.~\ref{fig:multi_long}, the input prompts focus on action changes, thus the generated videos should preserve consistent backgrounds while yielding rational and text-aligned action transformations. Although the videos inferred by VidRD and GenL are semantically consistent with the textual descriptions, they exhibit significant discontinuities in both the characters and the backgrounds. Under guidance from new prompts, FreeNoise induces unexpected changes within the scene, such as the man progressively transforming into a tree on the beach. MTVG exhibits noticeable changes in the background. In contrast, CoNo maintains temporal coherence across different video segments and achieves stable transitions due to our two proposed innovations: the beach background is preserved, and the same man performs corresponding actions.

For quantitative assessment, we mostly rely on the test set from MTVG to randomly sample $20$ videos per scenario. Tab.~\ref{tab:multi_long} indicates that CoNo and FreeNoise perform similarly on the CLIP-Text score, both outperforming other models. Local-window based FreeNoise excels in the CLIP-Image score (measuring the cosine similarity between two consecutive frames) due to its slow changes between adjacent frames. However, since the current scene may gradually transition to a new scene over the long term, we still assess Content Consistency (Temporal) through Human Evaluation. Besides the three dimensions mentioned in Sec.~\ref{sec:single-prompt}, our user study further incorporates Realism inspired by MTVG, focusing primarily on evaluating the realism of the generated video in terms of background and object consistency. It is noticed that MTVG is not open source, so we opt to perform human evaluations using few video examples from its project page. We observe that CoNo achieves the highest scores on most criteria, with ratings from participants on a Likert scale ranging from $1$ to $5$.
\subsection{Ablation Studies}
\label{sec:ablation}
\begin{figure}[!h]
    \centering
    \includegraphics[width=0.99\linewidth]
    {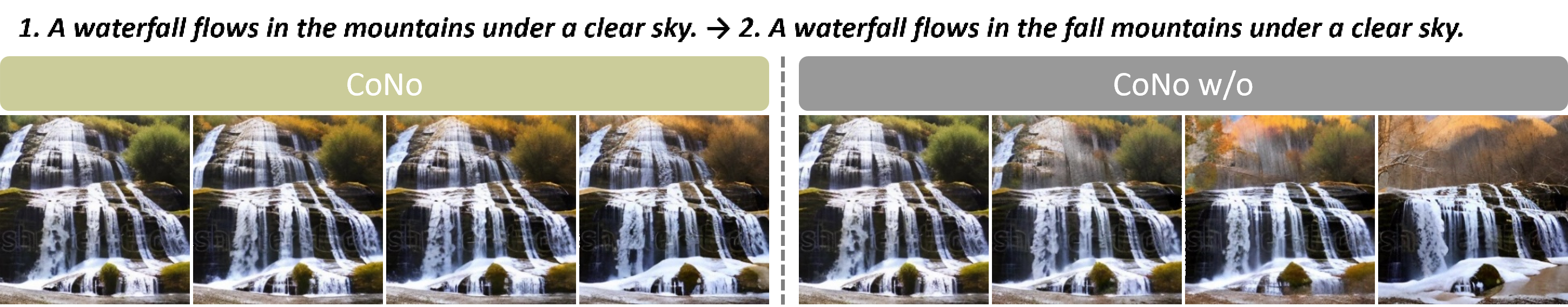}
    \caption{Ablation for Long-term Consistency Regularization. ``w/o'' indicates without the regularization in the CoNo pipeline.}
    \label{fig:AblationForRegu}
\end{figure}

\begin{figure}[!h]
    \centering
    \includegraphics[width=0.99\linewidth]
    {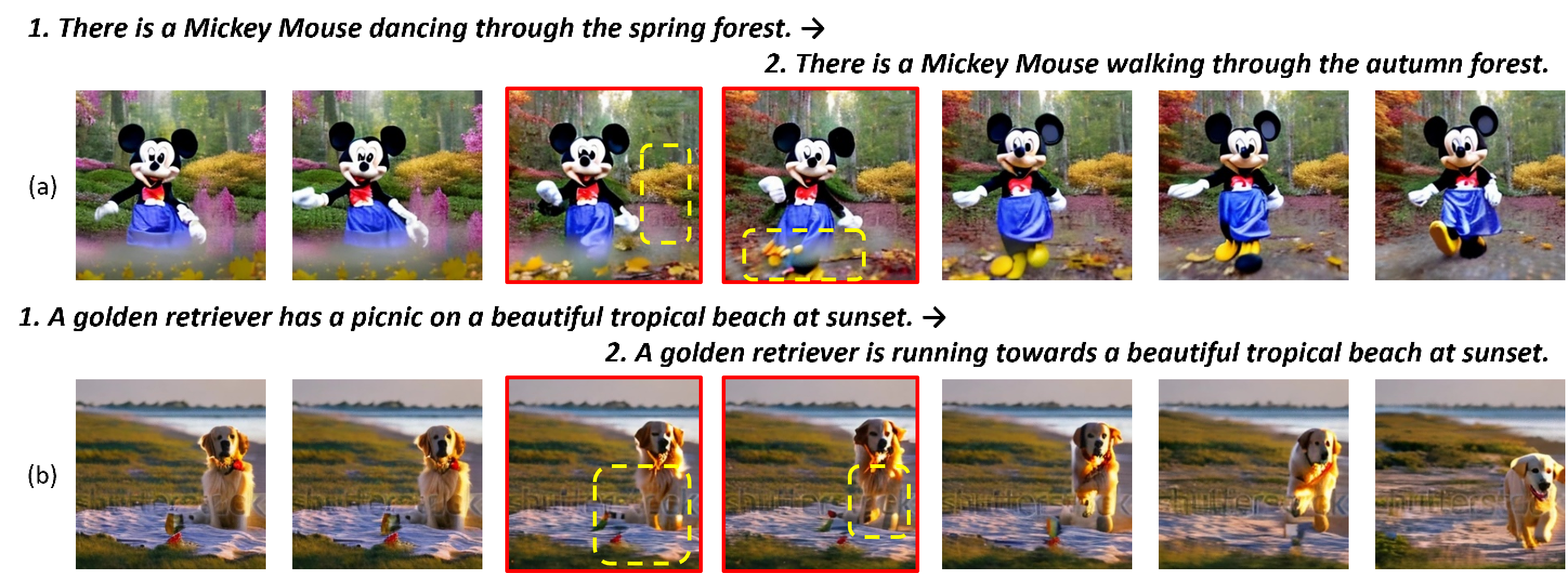}
    \caption{Ablation for Internal Noise Prediction. Transition frames are marked with red boxes and the details of transitions are highlighted with yellow boxes.}
    \label{fig:AblationForInter}
\end{figure}

\paragraph{Ablation for Long-term Consistency Regularization.} We qualitatively demonstrate the enhancement of scene consistency by long-term consistency regularization, using identical multi-text prompts and the same random seed for video generation, as shown in Fig.~\ref{fig:AblationForRegu}. While the ``look-back'' mechanism allows video frames to evolve semantically with the text and retain partial content, the divergence from the initial frame increases over time. Conversely, long-term consistency regularization preserves content consistency between frames.
\vspace{-5mm}
\paragraph{Ablation for Internal Noise Prediction.} In our proposed ``look-back'' mechanism, we introduce an internal noise prediction stage to facilitate appropriate transitions between different scenes. Fig.~\ref{fig:AblationForInter} illustrates this with two examples, highlighting the transition frames in red boxes. In Fig.~\ref{fig:AblationForInter} (a), the transition frames depict changes in both the environment and the foot movements of Mickey Mouse aligned with the text semantics. In Fig.~\ref{fig:AblationForInter} (b), a golden retriever is shown gradually transitioning from a sitting to a standing position. When the internal noise prediction stage is omitted, while content consistency is maintained, the transitions of scenes appear more abrupt.

To demonstrate the rationale for using the same set of initial noise for different video clips, we conducted an additional ablation study, as detailed in Sec.~\ref{sec:additional ablation studies}.
\section{Conclusion}
In this paper, we propose the CoNo, a novel tuning-free video diffusion for the generation of extended videos. This model incorporates two primary components: the ``look-back'' mechanism and the long-term consistency regularization. The ``look-back'' mechanism integrates an internal noise prediction stage within two video extending stages to enhance scene consistency across various video clips. Meanwhile, long-term consistency regularization addresses content shifts by capturing and constraining the global temporal information of adjacent video clips. Collectively, these innovations enable CoNo to effectively generate longer videos under both single-text and multi-text conditions.


\bibliography{main}
\bibliographystyle{abbrv}


\clearpage
\newpage
\appendix
\section{Appendix / supplemental material}
In Sec.~\ref{sec: Additional Implementation Details}, we provide additional implementation details, including the video diffusion models involved in this paper, the hyperparameter settings for CoNo, and the details of quantitative metrics and the user study. In Sec.~\ref{sec: Generalization Validation}, we validate the generalization of CoNo and present the qualitative results after replacing the base model with Lavie. In Sec.~\ref{sec: extended experiments}, we briefly explore the enhancements in long video continuity brought by prompt engineering. In Sec.~\ref{sec:additional ablation studies}, we provide the additional ablation study for the initial noise. Sec.~\ref{sec: More Qualitative Results} presents more qualitative results under single-text and multi-text prompt conditions. In Secs.~\ref{sec: Broader Impact} and ~\ref{sec: Limitations}, we discuss the broader impact and limitations, respectively. We include the pseudo-code of CoNo in Sec.~\ref{sec:pseudo-code} to facilitate reproduction.
\subsection{Additional Implementation Details}
\label{sec: Additional Implementation Details}
\paragraph{Video Diffusion Models.} Here, we list all the open-source video generation models involved in this paper for reproducibility:
\begin{itemize}
    \item VideoCrafter1~\cite{chen2023videocrafter1}: \href{https://github.com/AILab-CVC/VideoCrafter}{https://github.com/AILab-CVC/VideoCrafter}
    \item LVDM~\cite{he2022lvdm}: \href{https://github.com/YingqingHe/LVDM}{https://github.com/YingqingHe/LVDM}
    \item Gen-L-Video~\cite{wang2023gen-l-video}: \href{https://github.com/G-U-N/Gen-L-Video}{https://github.com/G-U-N/Gen-L-Video}
    \item VidRD~\cite{gu2023ReuseAndDiffuse}: \href{https://github.com/anonymous0x233/ReuseAndDiffuse}{https://github.com/anonymous0x233/ReuseAndDiffuse}
    \item FreeNoise~\cite{qiu2023freenoise}: \href{https://github.com/AILab-CVC/FreeNoise}{https://github.com/AILab-CVC/FreeNoise}
    \item Lavie~\cite{wang2023lavie}: \href{https://github.com/Vchitect/LaVie}{https://github.com/Vchitect/LaVie}
\end{itemize}
When the base model is VideoCrafter1, the number of frames $N$ is set to $16$. Accordingly, we set the hyperparameters in CoNo as follows: $T_d$ to $10$, $\delta$ to $140$, $N_1$ to $6$, and $N_2$ to $8$. When the base model is changed to Lavie, most settings remain the same, except $\delta$ is set to $260$. During sampling, we implemented DDIM~\cite{song2020ddim} with $50$ denoising steps and set DDIM eta to $0$. The value of the classifier-free guidance was set to $15$.

We use $700$ texts from Evalcrafter~\cite{liu2023evalcrafter} and four random seeds to generate a total of $2800$ longer videos under the single-text prompt condition, each video comprising $58$ frames. We removed the transitional frames from each long video and split them into three segments of $16$ frames, resulting in a total of $8400$ short videos. The base models also generate the same number of short videos, which facilitates the subsequent calculation of metrics such as FVD and KVD. The test set used under the multi-text prompt condition can be found in MTVG~\cite{oh2023mtvg}.

For the user study, we adopt the five-point scale method from MTVG, asking $13$ participants to rate randomly shuffled videos on the following questions: (1) \textit{How smoothly does the content of the videos change in response to the given prompts? (Temporal Consistency)}; (2) \textit{How well does the video correspond to the prompts? (Semantic Alignment)}; (3) \textit{How natural and realistic does this video look, considering the consistency of the background and the objects? (Realism)}; (4) \textit{Considering the three questions above, please rank the overall video quality. (Preference)}. These four questions are used under the multi-text prompt condition, while under the single-text prompt condition, similar questions are used but without the Realism aspect. We randomly selected $60$ single-text prompts from Evalcrafter and $30$ sets of multi-text prompts provided by MTVG to generate videos for participant evaluation.
\subsection{Generalization Validation}
\label{sec: Generalization Validation}
\begin{figure}[!h]
    \centering
    \includegraphics[width=0.99\linewidth]
    {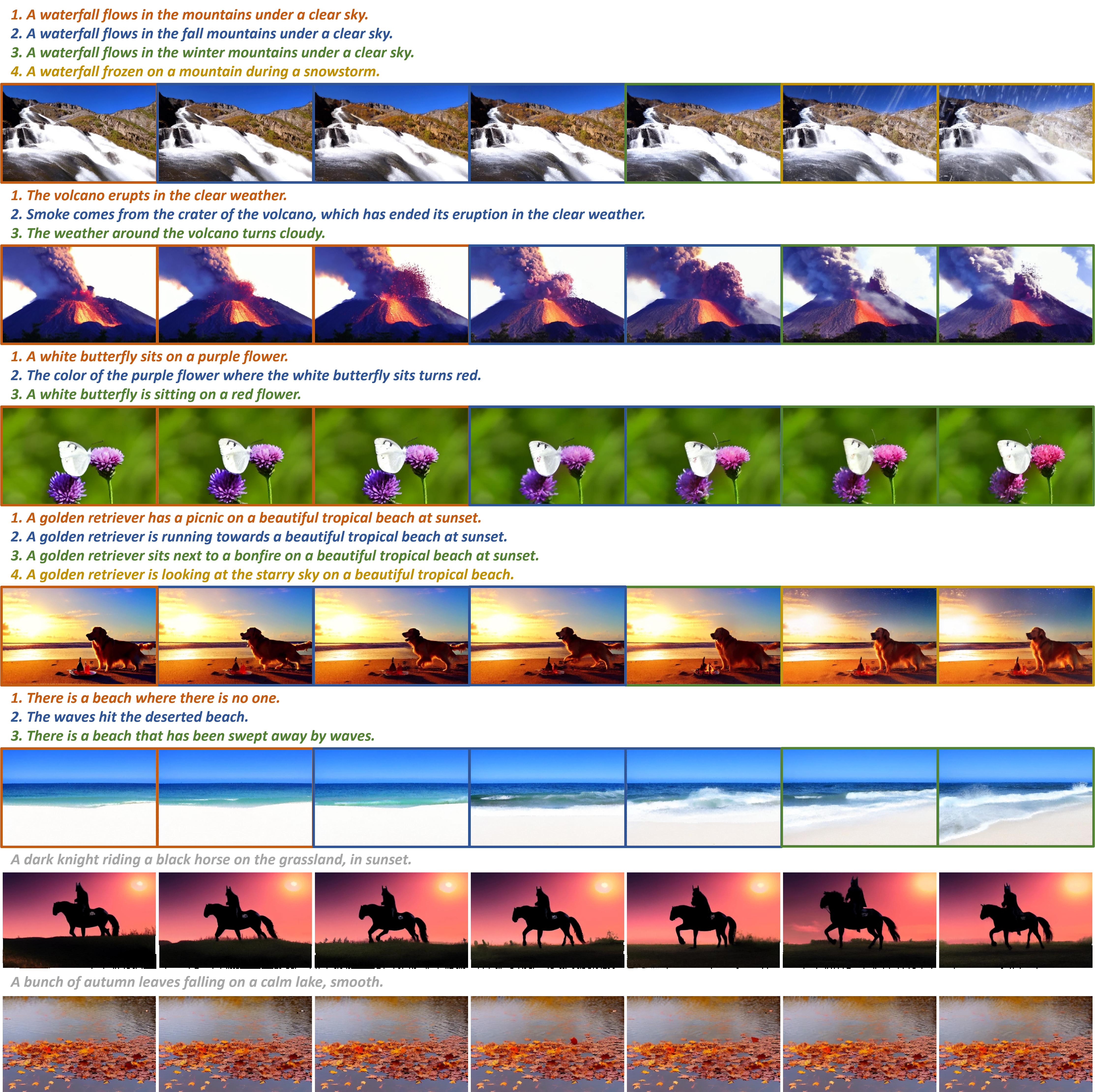}
    \caption{Generalization validation with the base model replaced by Lavie.}
    \label{fig:lavie valid}
\end{figure}

To validate the generalization of the current CoNo approach, we replace the base model VideoCrafter1 with Lavie~\cite{wang2023lavie} and present the results in Fig.~\ref{fig:lavie valid}. It can be observed that our strategy remains effective under both single-text and multi-text prompt conditions.
\subsection{Expansion Experiments}
\label{sec: extended experiments}
\begin{figure}[!h]
    \centering
    \includegraphics[width=0.99\linewidth]
    {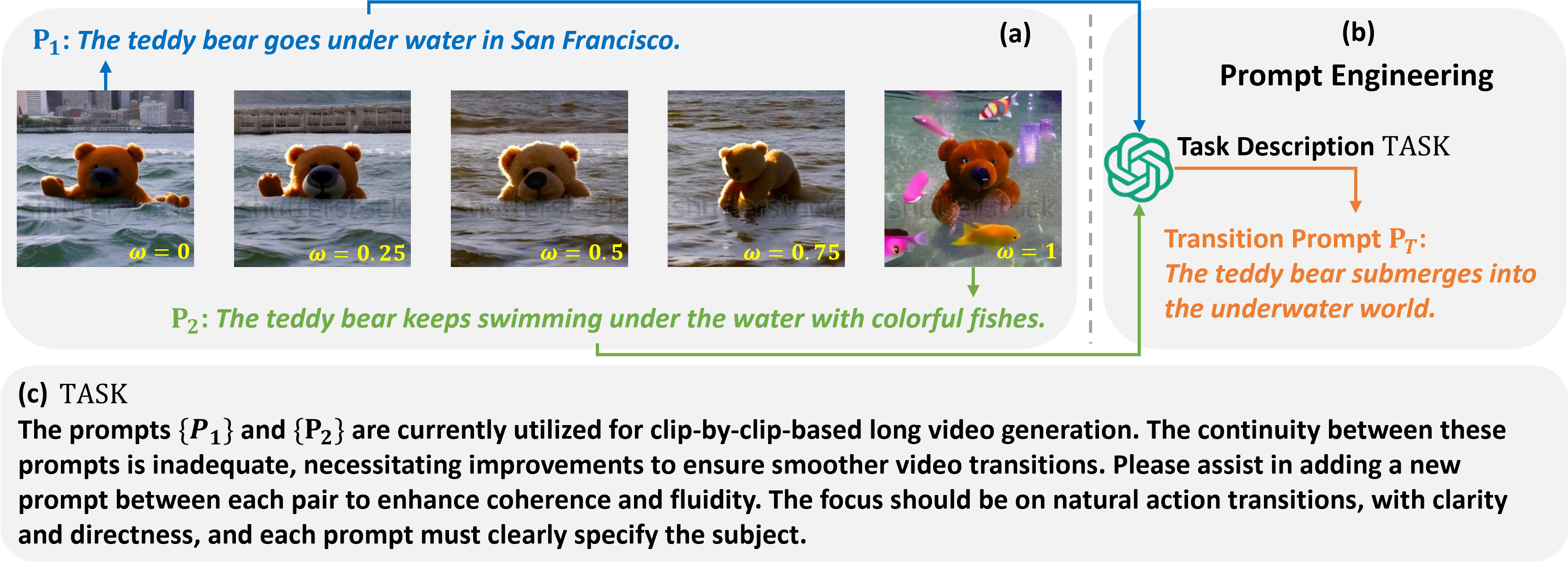}
    \caption{Using commonsense knowledge from GPT-4 for prompt engineering to facilitate more logical scene transitions by generating semantically transitional text prompts.}
    \label{fig:PromptInterpolation}
\end{figure}
\begin{figure}[!h]
    \centering
    \includegraphics[width=0.99\linewidth]
    {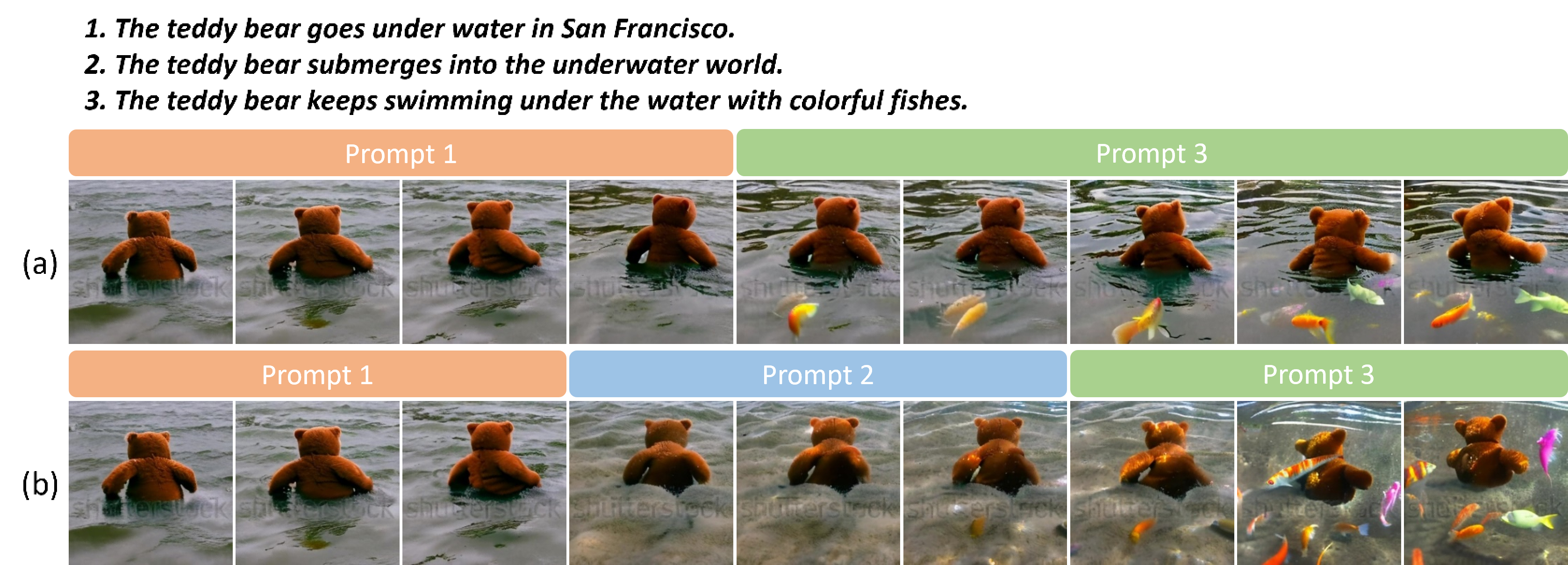}
    \caption{Comparison Results of prompt engineering}
    \label{fig:vs prompt}
\end{figure}
\paragraph{Enhancing Video Continuity through Prompt Engineering.} Prompts play a vital role in facilitating interaction in Text-to-Video generation~\cite{wang2024VidProM}, and to enhance the continuity of extended videos, we explore prompt engineering in this section. 
FreeNoise~\cite{qiu2023freenoise} utilizes a linear interpolation of prompt embeddings to facilitate smooth transitions. Yet, our experimental findings indicate that the motions and scenes in videos generated from these intermediate embeddings may not consistently exhibit transitional characteristics.
We apply linear interpolation between two distinct text embeddings, $e_1$ and $e_2$, following the formula $e_{\mathrm{lerp}} = e_1 + \omega (e_2 - e_1)$. By adjusting the interpolation weight $\omega$ and maintaining the same random seed for initialization, we successfully generate a series of videos depicted in Fig.~\ref{fig:PromptInterpolation} (a). We observe that prompt embedding interpolation exhibits limited continuity in semantic transitions, reminding us to utilize the common sense of GPT-4~\cite{achiam2023gpt4} to generate smoother transitional prompts between adjacent text prompts. As shown in Fig.~\ref{fig:PromptInterpolation} (b), upon inputting an appropriate task description $\mathrm{TASK}$ along with texts $P_1$ and $P_2$ into GPT-4, the model is enabled to infer an intermediary text $P_T$ that semantically bridges $P_1$ and $P_2$ for video expansion, which ensures a more natural transition of object movements and background changes in the video. 

We show the improvements in video continuity afforded by prompt engineering in Fig.~\ref{fig:vs prompt}. The first and third text prompts in Fig.~\ref{fig:vs prompt} are taken from the original test set in MTVG. Fig.~\ref{fig:vs prompt} (a) shows the videos generated under these two prompts. It can be observed that the teddy bear does not transition to ``under the water'' in the scene ``The teddy bear keeps swimming under the water with colorful fishes.'' To address this, we use GPT-4 to generate a new prompt, ``The teddy bear submerges into the underwater world,'' which logically facilitates the scene transition, and place it between the existing two prompts. Compared to embedding interpolation, this method of prompt interpolation serves more as a semantic transition. After applying prompt engineering, the video shown in Fig.~\ref{fig:vs prompt} (b) better matches the scene description.
\subsection{Additional Ablation Study}
\label{sec:additional ablation studies}
\begin{figure}[!h]
    \centering
    \includegraphics[width=0.99\linewidth]
    {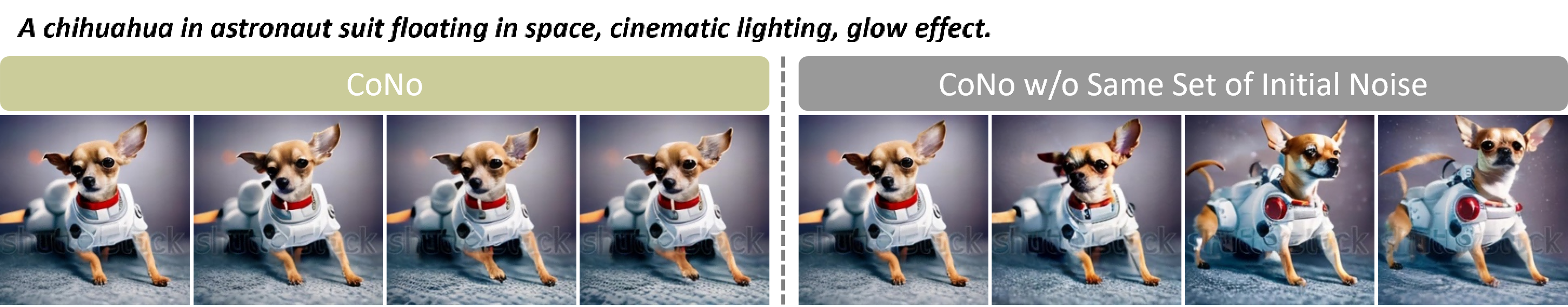}
    \caption{Ablation for Initial Noise. ``w/o'' indicates without the same set of initial noise.}
    \label{fig:AblationForNoise}
\end{figure}
\paragraph{Ablation for Initial Noise.} To verify that maintaining the same set of initial noise (even in a different order) is crucial for the consistency of the final generated content, we replace the initial noise of one non-guided frame with another randomly sampled Gaussian noise during the video extending stage. The comparison results are shown in the Fig.~\ref{fig:AblationForNoise}. It is evident that even one changed noise frame in the initial noise significantly damages the consistency of the long video. The chihuahua's astronaut suit and the space environment behind it have both altered.
\subsection{More Qualitative Results}
\label{sec: More Qualitative Results}
\begin{figure}[!h]
    \centering
    \includegraphics[width=0.99\linewidth]
    {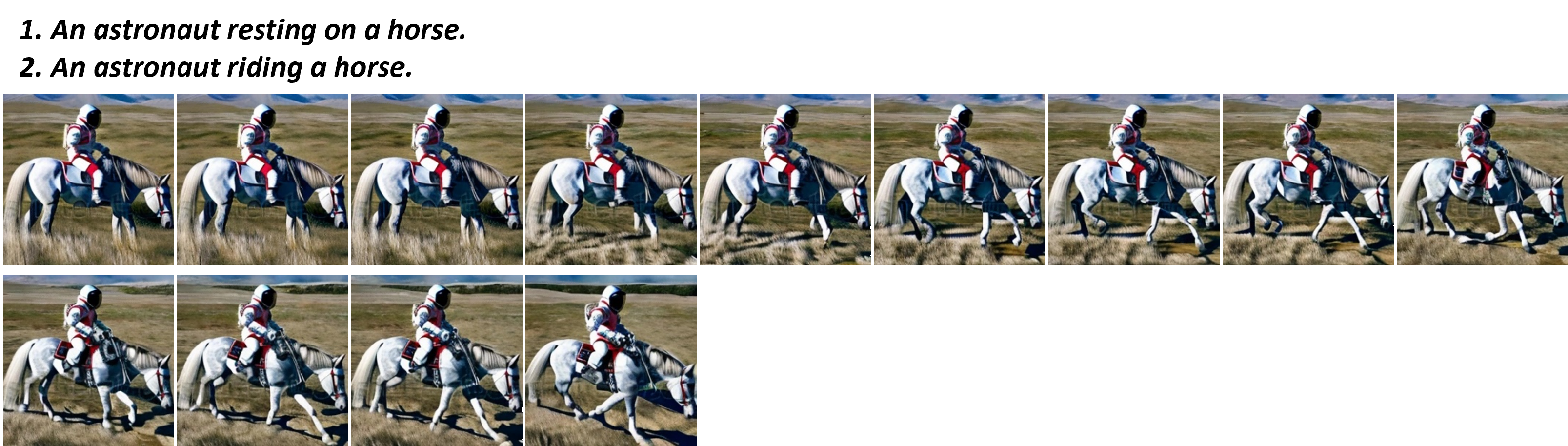}
    \caption{Additional qualitative results conditioned on multi-text with VideoCrafter1.}
    \label{fig:multi horse}
\end{figure}
\begin{figure}[!h]
    \centering
    \includegraphics[width=0.99\linewidth]
    {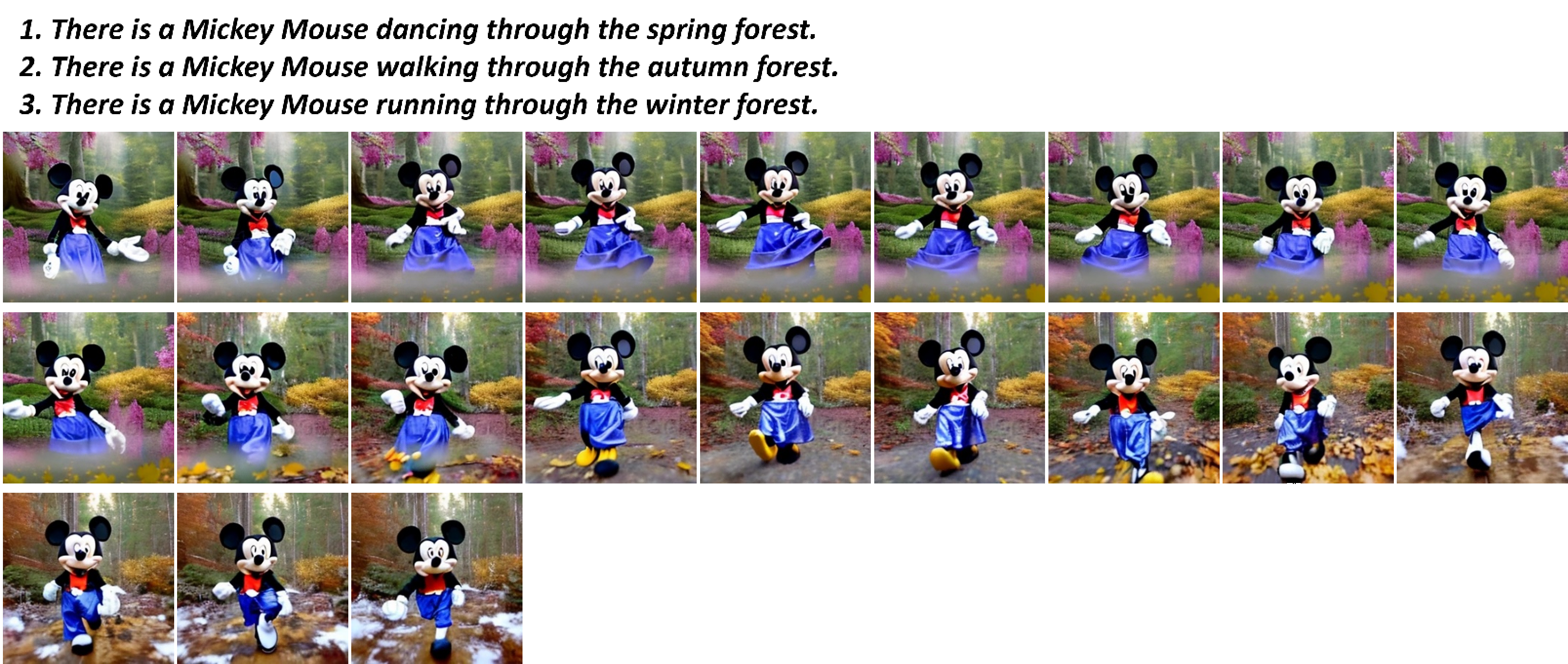}
    \caption{Additional qualitative results conditioned on multi-text with VideoCrafter1.}
    \label{fig:multi mouse}
\end{figure}
\begin{figure}[!h]
    \centering
    \includegraphics[width=0.99\linewidth]
    {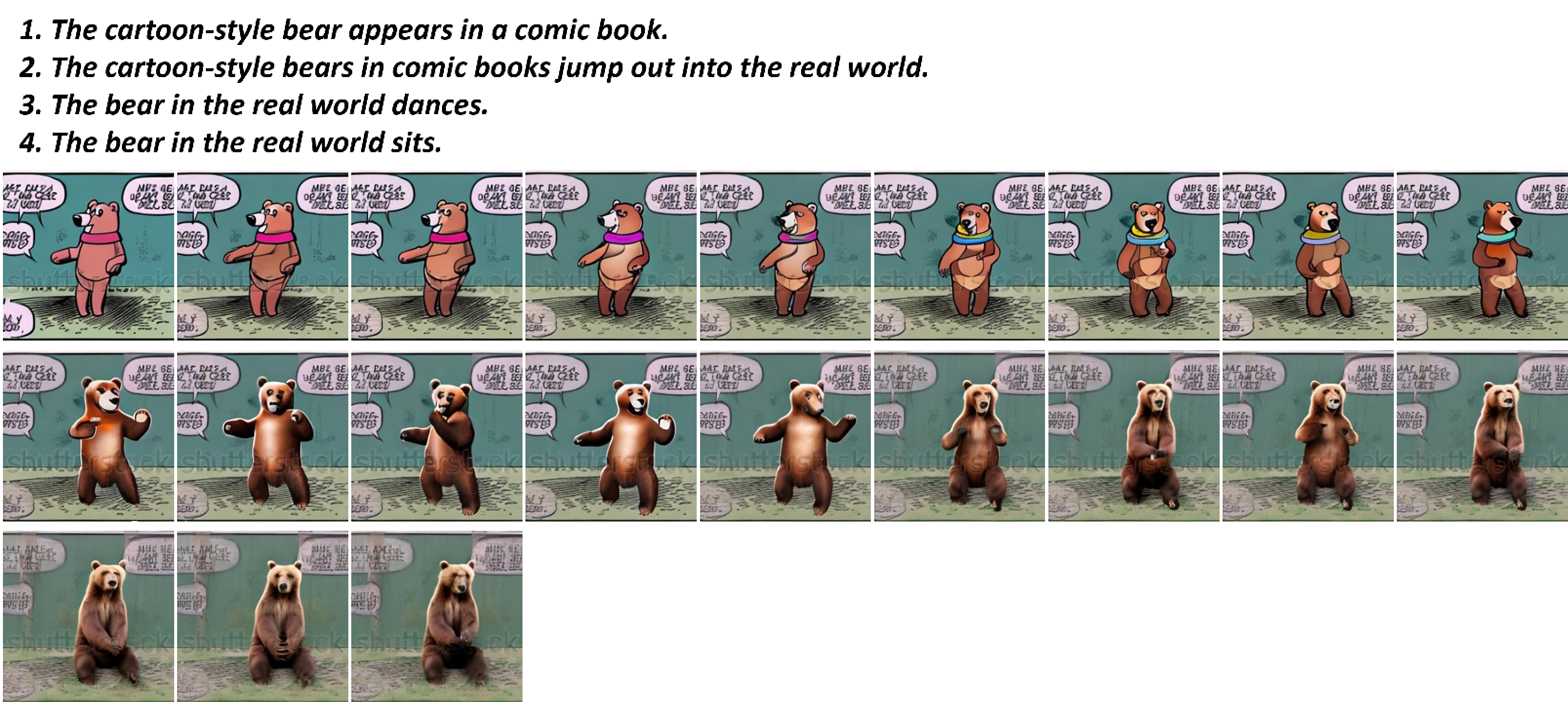}
    \caption{Additional qualitative results conditioned on multi-text with VideoCrafter1.}
    \label{fig:multi bear}
\end{figure}
\begin{figure}[!h]
    \centering
    \includegraphics[width=0.99\linewidth]
    {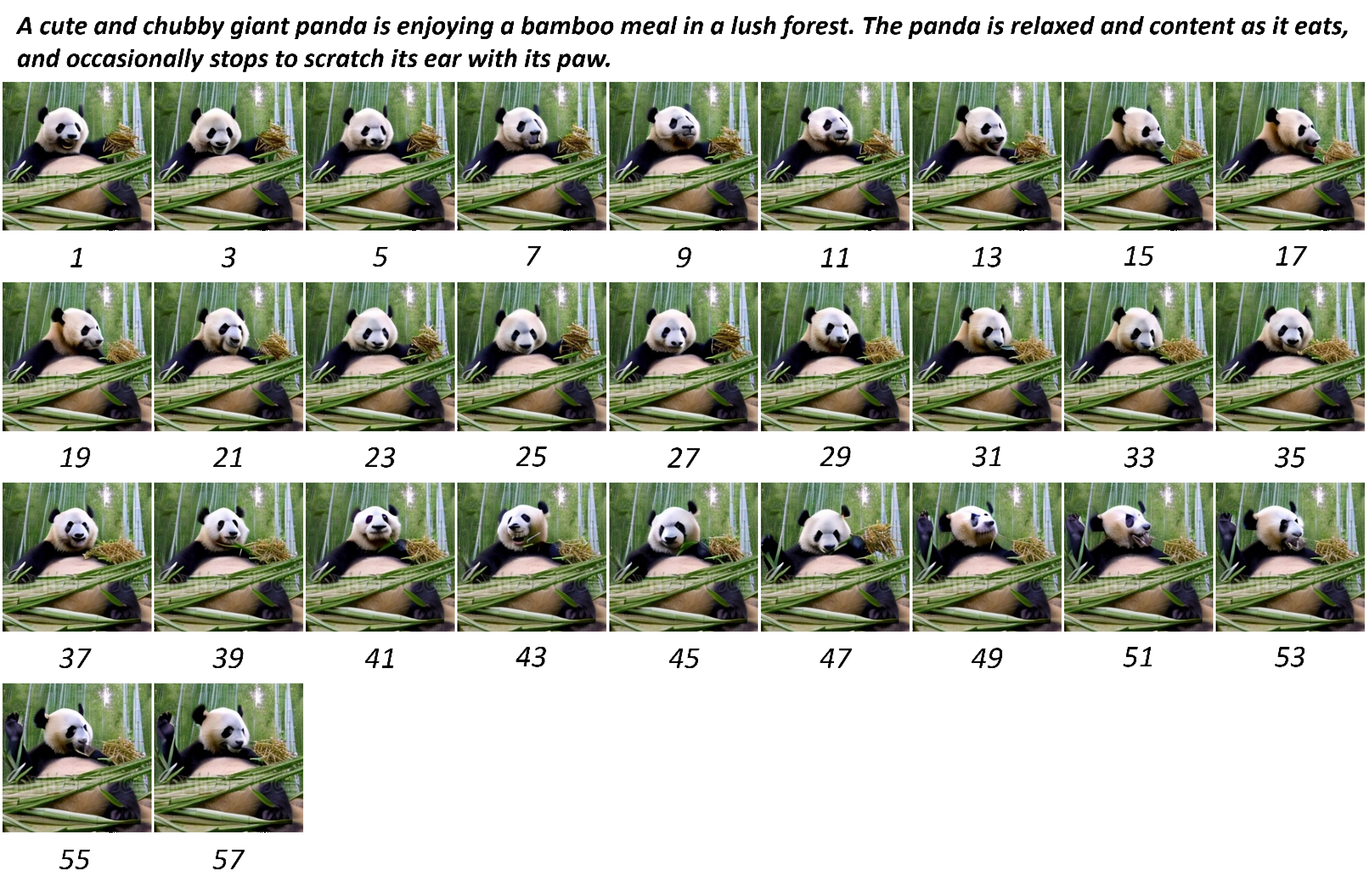}
    \caption{Additional qualitative results conditioned on single-text with VideoCrafter1.}
    \label{fig:single panda}
\end{figure}
\begin{figure}[!h]
    \centering
    \includegraphics[width=0.99\linewidth]
    {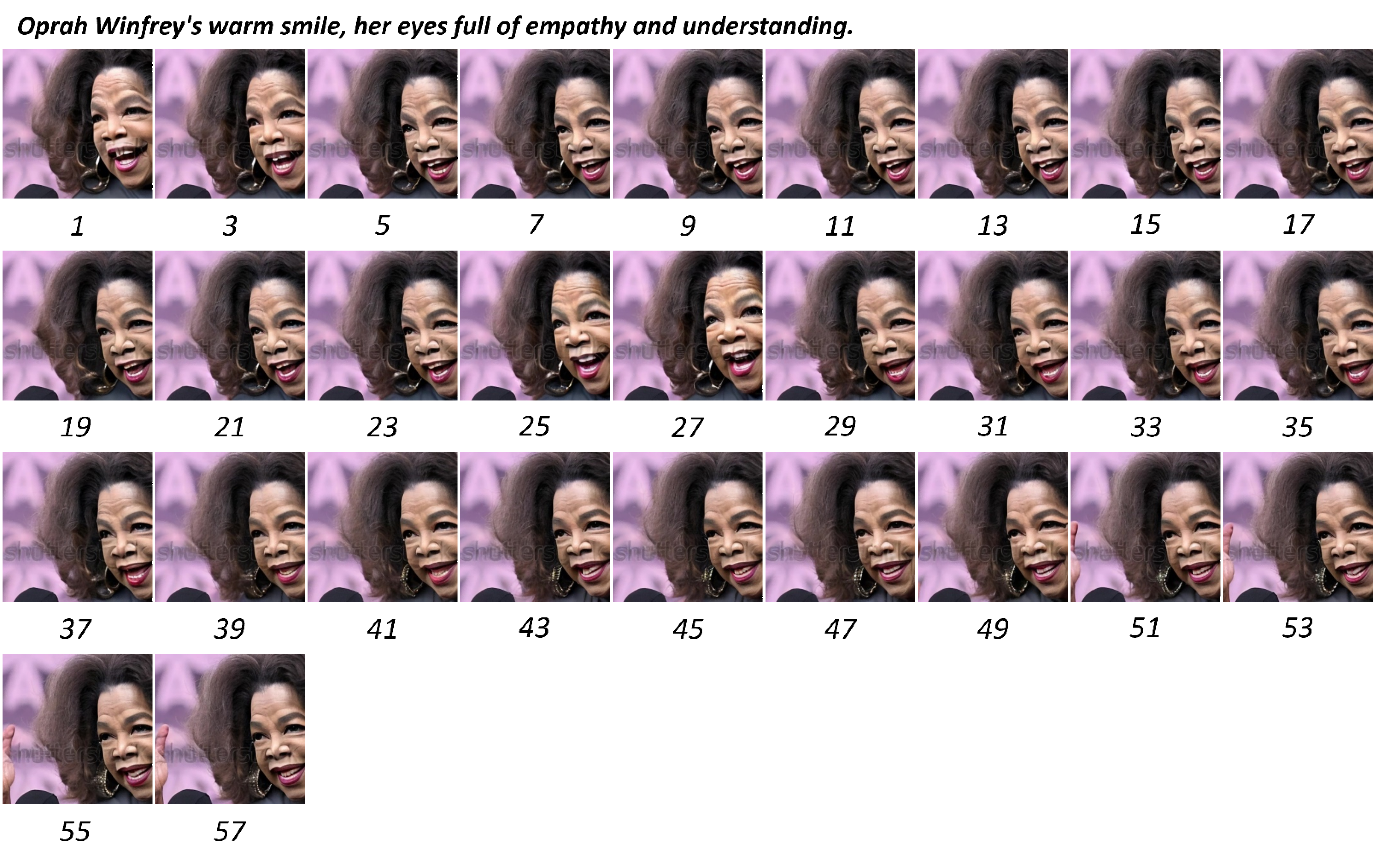}
    \caption{Additional qualitative results conditioned on single-text with VideoCrafter1.}
    \label{fig:single person}
\end{figure}

In this section, we provide more qualitative results. Figs.~\ref{fig:single panda} and ~\ref{fig:single person} show results for single-text inputs, while Figs.~\ref{fig:multi horse}, ~\ref{fig:multi mouse}, and ~\ref{fig:multi bear} show results for multi-text inputs.The base model used in all cases is VideoCrafter1. Frame numbers are annotated below some images.
\subsection{Broader Impact}
\label{sec: Broader Impact}
Our work on tuning-free long video generation using existing short video diffusion models has several potential impacts. It offers a resource-friendly solution for long video generation, making it accessible to a wider range of users. In addition, users can create high-quality content by utilizing different base models that meet their specific needs. Academically, this approach encourages further research in enhancing video consistency and generation efficiency. However, it is crucial to consider ethical implications, such as the potential for misuse in creating misleading or harmful content. We emphasize the importance of developing and adhering to responsible usage guidelines to mitigate these risks.
\subsection{Limitations}
\label{sec: Limitations}
While our tuning-free approach to long video generation shows promise, there is a potential limitation caused by the pre-trained video diffusion models. Limited by the current state of base video generation models, they might generate imperfect results when users input complex texts, such as specifying the number of objects in the scene or their relative positions in some cases. However, considering that base models have been rapidly developing and our CoNo applies to various base models, this issue is expected to be resolved, thereby enhancing the effectiveness of our approach.
\subsection{Pseudo-Code of CoNo}
\label{sec:pseudo-code}
\begin{algorithm}
\caption{Pseudo-Code of CoNo}
\begin{algorithmic}[1]
\Require A sequence of event prompts $\left(P_1, P_2, P_3, \cdots\right)$ and the base video diffusion model (\textit{e.g.}, VideoCrafter1)
\Ensure Extended video $z_{Final}$ with scene consistency
\footnotesize 
\State Initialize the base model with pre-trained weights and randomly sample noise frames $z_{T}^{P_1}$. Generate the first video clip $z_{0}^{P_1}$ under the guidance of $P_{1}$ and store the predicted noise $\hat{\epsilon}_{t}^{P_1}$ of each timestep $t$. $z_{Final} = z_{0}^{P_1}$
\For{each pair of adjacent event prompts (\textit{e.g.}, $P_{1}$, $P_{2}$)}
    \If{is the first video extending stage}
        \State $z_{T}^{P_2} = \text{torch.flip}(z_{T}^{P_1}, \text{dims}=[2])$
        \State $z_{T}^{P_2}\left[ :,:,0:N_1,:,: \right] = \text{torch.flip}(z_{T}^{P_1}\left[ :,:,0:N_1,:,: \right], \text{dims}=[2])$
        \For{each timestep $t$}
            \State Predict noise $\hat{\epsilon}_{t}^{P_2}$ using the base model under the guidance of $P_{2}$ and store $\hat{\epsilon}_{t}^{P_2}$
            \State Compute $\hat{\epsilon}_{t,content}^{P_1} = \left( \sum_{n=0}^{N-1}{\hat{\epsilon}_{t}^{P_1}} \right) / N$
            \State Compute $\hat{\epsilon}_{t,content}^{P_2} = \left( \sum_{n=0}^{N-1}{\hat{\epsilon}_{t}^{P_2}} \right) / N$
            \State Update $\hat{\epsilon}_{t}^{P_2} \gets \hat{\epsilon}_{t}^{P_2} - \delta \nabla _{\hat{\epsilon}_{t}^{P_2}} g\left( \hat{\epsilon}_{t,content}^{P_1}, \hat{\epsilon}_{t,content}^{P_2} \right)$
            \If{$t \geq T_d$}
                \State $\hat{\epsilon}_{t}^{P_2}\left[ :,:,0:N_1,:,: \right] = \hat{\epsilon}_{t}^{P_1}\left[ :,:,0:N_1,:,: \right]$
            \EndIf
        \EndFor
        \State Obtain $z_{0}^{P_2}$
    \EndIf
    \If{is the internal noise prediction stage}
        \State $z_{T}^{P_{2^{\prime}}} = \text{torch.cat}([z_{T}^{P_2}\left[ :,:,0:N_1,:,: \right], z_{T}^{P_2}\left[ :,:,N_1+N_2:N,:,: \right], z_{T}^{P_2}\left[ :,:,N_1:N_1+N_2,:,: \right]], \text{dim}=2)$
        \For{each timestep $t$}
            \State Predict noise $\hat{\epsilon}_{t}^{P_{2^{\prime}}}$ using the base model under the guidance of $P_{2}$ and store $\hat{\epsilon}_{t}^{P_{2^{\prime}}}$
            \State Compute $\hat{\epsilon}_{t,content}^{P_2} = \left( \sum_{n=0}^{N-1}{\hat{\epsilon}_{t}^{P_2}} \right) / N$
            \State Compute $\hat{\epsilon}_{t,content}^{P_{2^{\prime}}} = \left( \sum_{n=0}^{N-1}{\hat{\epsilon}_{t}^{P_{2^{\prime}}}} \right) / N$
            \State Update $\hat{\epsilon}_{t}^{P_{2^{\prime}}} \gets \hat{\epsilon}_{t}^{P_{2^{\prime}}} - \delta \nabla _{\hat{\epsilon}_{t}^{P_{2^{\prime}}}} g\left( \hat{\epsilon}_{t,content}^{P_2}, \hat{\epsilon}_{t,content}^{P_{2^{\prime}}} \right)$
            \If{$t \geq T_d$}
                \State $\hat{\epsilon}_{t}^{P_{2^{\prime}}}\left[ :,:,0:N_1,:,: \right] = \hat{\epsilon}_{t}^{P_2}\left[ :,:,0:N_1,:,: \right]$
                \State $\hat{\epsilon}_{t}^{P_{2^{\prime}}}\left[ :,:,N-N_2:N,:,: \right] = \hat{\epsilon}_{t}^{P_2}\left[ :,:,N-N_2:N,:,: \right]$
            \EndIf
        \EndFor
        \State Obtain $z_{0}^{P_{2^{\prime}}}$
    \EndIf
    \If{is the second video extending stage}
        \State $z_{T}^{P_{2^{\prime\prime}}} = \text{torch.flip}(z_{T}^{P_{2^{\prime}}}, \text{dims}=[2])$
        \State $z_{T}^{P_{2^{\prime\prime}}}\left[ :,:,0:N_1,:,: \right] = \text{torch.flip}(z_{T}^{P_{2^{\prime}}}\left[ :,:,0:N_1,:,: \right], \text{dims}=[2])$
        \For{each timestep $t$}
            \State Predict noise $\hat{\epsilon}_{t}^{P_{2^{\prime\prime}}}$ using the base model under the guidance of $P_{2}$ and store $\hat{\epsilon}_{t}^{P_{2^{\prime\prime}}}$
            \State Compute $\hat{\epsilon}_{t,content}^{P_{2^{\prime}}} = \left( \sum_{n=0}^{N-1}{\hat{\epsilon}_{t}^{P_{2^{\prime}}}} \right) / N$
            \State Compute $\hat{\epsilon}_{t,content}^{P_{2^{\prime\prime}}} = \left( \sum_{n=0}^{N-1}{\hat{\epsilon}_{t}^{P_{2^{\prime\prime}}}} \right) / N$
            \State Update $\hat{\epsilon}_{t}^{P_{2^{\prime\prime}}} \gets \hat{\epsilon}_{t}^{P_{2^{\prime\prime}}} - \delta \nabla _{\hat{\epsilon}_{t}^{P_{2^{\prime\prime}}}} g\left( \hat{\epsilon}_{t,content}^{P_{2^{\prime}}}, \hat{\epsilon}_{t,content}^{P_{2^{\prime\prime}}} \right)$
            \If{$t \geq T_d$}
                \State $\hat{\epsilon}_{t}^{P_{2^{\prime\prime}}}\left[ :,:,0:N_1,:,: \right] = \hat{\epsilon}_{t}^{P_{2^{\prime}}}\left[ :,:,0:N_1,:,: \right]$
            \EndIf
        \EndFor
        \State Obtain $z_{0}^{P_{2^{\prime\prime}}}$
    \EndIf
    \State $z_{Final}$.append($z_{0}^{P_{2^{\prime}}}\left[ :,:,N_1:N-N_1,:,: \right], z_{0}^{P_{2^{\prime\prime}}}$)
\EndFor
\State Obtain $z_{Final}$
\normalsize 
\end{algorithmic}
\end{algorithm}

\end{document}